\documentclass[letterpaper, 10 pt, journal, twoside]{IEEEtran}
\IEEEoverridecommandlockouts
\usepackage[final]{changes}
\usepackage{amsmath,amsfonts}
\usepackage{algorithmic}
\usepackage{algorithm}
\usepackage{array}
\usepackage[caption=false,font=normalsize,labelfont=sf,textfont=sf]{subfig}
\usepackage{textcomp}
\usepackage{stfloats}
\usepackage{url}
\usepackage{verbatim}
\usepackage{graphicx}
\usepackage{cite}
\usepackage{threeparttable}
\usepackage{bm}
\usepackage{hyperref}
\usepackage{booktabs}
\newcommand{\link}[1]{{\color{blue}\href{#1}{#1}}}
\usepackage{color, colortbl,soul}
\setdeletedmarkup{\color{red}\sout{#1}}

\usepackage{eso-pic}
\usepackage{url}
\AddToShipoutPicture*{%
     \AtTextUpperLeft{%
         \put(-3.5,10){
           \begin{minipage}{\textwidth}
              \scriptsize
              \MakeUppercase{Final version available at} \url{https://doi.org/10.1109/LRA.2023.3290420}
           \end{minipage}}%
     }%
}

\begin{document}

\title{Predicting Class Distribution Shift for Reliable Domain Adaptive Object Detection}
\author{Nicolas Harvey Chapman$^{1}$, Feras Dayoub$^{2}$, Will Browne$^{1}$ and Christopher Lehnert$^{1}$\vspace{-2em}%
\thanks{Manuscript received: February 5, 2023; Revised May 8, 2023; Accepted June 14, 2023. This paper was recommended for publication by Editor Markus Vincze upon evaluation of the Associate Editor and Reviewers' comments.}%
\thanks{$^{1}$Nicolas Harvey Chapman, Will Browne and Christopher Lehnert are with the School of Electrical Engineering and Robotics, Queensland University of Technology, Brisbane, Australia ({\tt\footnotesize will.browne@qut.edu.au; c.lehnert@qut.edu.au; nicolasharvey.chapman@hdr.qut.edu.au}).}%
\thanks{$^{2} $Feras Dayoub is with the School of Computer Science and the Australian Institute of Machine Learning at the University of Adelaide, Adelaide, Australia ({\tt\footnotesize feras.dayoub@adelaide.edu.au}).}%
\thanks{Digital Object Identifier (DOI): see top of this page.}
}

\markboth{IEEE Robotics and Automation Letters. Preprint Version. Accepted June, 2023}
{Chapman \MakeLowercase{\textit{et al.}}: Predicting Class Distribution Shift for Reliable
Domain Adaptive Object Detection}

\maketitle
\begin{abstract}
Unsupervised Domain Adaptive Object Detection (UDA-OD) uses unlabelled data to improve the reliability of robotic vision systems in open-world environments. Previous approaches to UDA-OD based on self-training have been effective in overcoming changes in the general appearance of images. However, shifts in a robot's deployment environment can also impact the likelihood that different objects will occur, termed class distribution shift. Motivated by this, we propose a framework for explicitly addressing class distribution shift to improve pseudo-label reliability in self-training. Our approach uses the domain invariance and contextual understanding of a pre-trained joint vision and language model to predict the class distribution of unlabelled data. By aligning the class distribution of pseudo-labels with this prediction, we provide weak supervision of pseudo-label accuracy. To further account for low quality pseudo-labels early in self-training, we propose an approach to dynamically adjust the number of pseudo-labels per image based on model confidence. Our method outperforms state-of-the-art approaches on several benchmarks, including a 4.7 mAP improvement when facing challenging class distribution shift. Code available at \link{https://github.com/nhcha6/ClassDistributionPrediction}

\end{abstract}

\begin{IEEEkeywords}
Object Detection, Deep Learning for Visual Perception; Visual Learning
\vspace{-1em}
\end{IEEEkeywords}

\section{Introduction}
\IEEEPARstart{O}{bject} detection is a crucial component of many robotic systems, from self-driving cars to service robots. Existing object detectors based on deep learning require the collection of large, annotated datasets for training. However, in open-world deployment a robot will encounter changes in object appearance due to factors such as weather, lighting conditions, or image corruptions \cite{dafrcnn}. Furthermore, shifts in a robot's deployment environment can impact the likelihood that different objects will occur, termed class distribution shift \cite{real_life_cl}. Due to the high costs of annotation, it is infeasible to gather labelled data for all potential conditions and environments \cite{on_the_job}. Thus, it is inevitable that an object detector deployed on a robot will face the problem of domain shift, where the images being processed do not match those used for training. Unfortunately, the performance of deep learning-based object detectors degrades significantly when facing such domain shift \cite{dafrcnn}. To address this issue, Unsupervised Domain Adaptive Object Detection (UDA-OD) has been proposed to adapt a model from a known source domain to a shifted target domain using only unlabelled data. This strategy can help the model generalise and improve its performance in the target domain, without the need for expensive labelling.

Self-training methods have recently produced State-Of-The-Art (SOTA) results on UDA-OD benchmarks \cite{prob_teacher, umt_uda}. Such approaches leverage the Mean Teacher framework \cite{mean_teacher} to enforce consistency between a student and teacher model on unlabelled images (Section III-B). Key to this framework is the use of confident detections from the teacher as pseudo-labels for training the student. These methods were initially used for Semi-Supervised Object Detection (SSOD) \cite{simple_ssod, unbiased1}, where the labelled and unlabelled data come from the same distribution. Thus, when applied to a challenging domain adaptation problem, the reliability of pseudo-labels generated by the teacher may degrade significantly \cite{umt_uda}. Focusing on this challenge, recent work has adapted Mean Teacher to UDA-OD by improving pseudo-label reliability in the presence of domain shift \cite{umt_uda, simrod, prob_teacher, object_relations, free_lunch, simrod}.

\begin{figure}[!t]
    \centering
    \includegraphics[width=0.99\columnwidth, height=0.20\textheight]{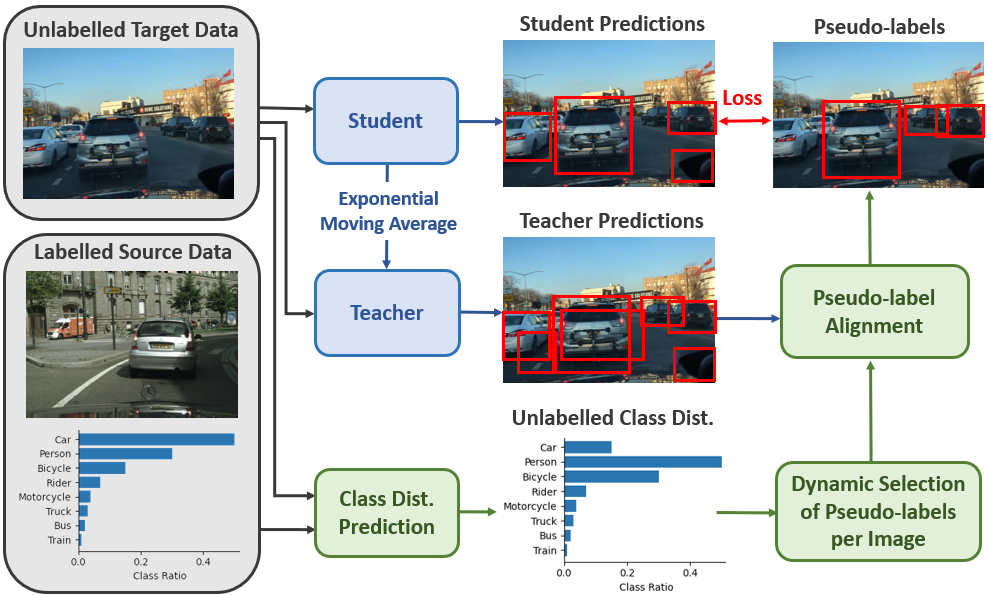}
    \vspace{-1.5em}
    \caption{Our framework for explicitly incorporating class distribution shift into self-training to improve pseudo-label reliability. As per the standard implementation of Mean Teacher, confident detections from a teacher model are used as pseudo-labels to train a student model using unlabelled data. The Exponential Moving Average (EMA) of the weights of the student are then used to update the teacher to make it more stable during training. Traditionally, a static confidence threshold is defined for all classes to generate pseudo-labels. Our method (shown in green) instead predicts the class distribution of the unlabelled data, and selects confidence thresholds to align the class distribution of the pseudo-labels with this prediction. To further address the poor performance of the teacher model in the target domain, we dynamically adjust the number of pseudo-labels per image as teacher confidence increases. 
    }
    \vspace{-1.5em}
    \label{fig:MethodOverView}
\end{figure}

While promising, existing UDA-OD methods and benchmarks focus largely on changes to the appearance of images \cite{dafrcnn}. Thus, the benefit of explicitly addressing class distribution shift during self-training is unexplored. Furthermore, in robotics there are opportunities to use contextual cues to generate a prediction for the likelihood of object occurrence. For example, one expects an autonomous vehicle to encounter more pedestrians on city streets than on the highway \cite{real_life_cl}. Motivated by this, we propose a framework for explicitly addressing class distribution shift during self-training. We find that our approach can significantly increase the reliability of pseudo-labels produced by Mean Teacher, leading to improved domain adaptation.

Our framework involves firstly predicting the class distribution of unlabelled data in novel deployment environments. Due to their strong understanding of image context and resistance to domain shift, we find that pre-trained joint vision and language models are useful for this task \cite{clip}. We then use Adapative label distribution aware Confidence Thresholding (ACT) \cite{labelmatch} (Section III-C) to incorporate our prediction into Mean Teacher. Proposed for SSOD, ACT improves performance on low probability classes and provides weak supervision of the teacher model by aligning the distribution of pseudo-labels with that of the labelled data. By using our prediction instead of the labelled prior, the reliability of pseudo-labels used in Mean Teacher improves substantially. However, even with a perfectly accurate class distribution, we find that the pseudo-labels generated with ACT remain unreliable early in self-training. To address this, we propose an approach for dynamically adjusting the number of pseudo-labels per image based on the confidence of the teacher model. Extensive experiments show that our proposed method returns SOTA performance on several benchmarks, including a 4.3 mAP improvement when adapting from a small to large scale dataset. On a novel scenario containing more challenging class distribution shift, we return a 4.7 mAP improvement.

To summarise, this letter makes the following contributions:
\begin{itemize}
    \item A framework for explicitly addressing class distribution shift during self-training, leading to an improvement in pseudo-label reliability.
    \item A method to predict the class distribution of unlabelled data using a pre-trained joint vision and language model.
    \item An approach for dynamically adjusting the number of pseudo-labels per image to account for the confidence of the teacher in the target domain.
    \item Experimental results showing that the proposed method returns SOTA performance on several benchmarks, including scenarios with realistic class distribution shift.
\end{itemize}
\vspace{-0.5em}
\section{Related Work}
\subsection{Self-training in Unsupervised Domain Adaptive Object Detection}
Initially proposed for semi-supervised learning \cite{mean_teacher}, self-training methods have recently produced SOTA results on UDA-OD benchmarks \cite{prob_teacher, umt_uda}. Several
augmentations have been made to the standard Mean Teacher framework \cite{mean_teacher} to improve pseudo-label reliability under domain shift. Methods have been proposed for merging patches from labelled and unlabelled images \cite{simrod}, selecting optimal pseudo-labels to balance true positive and false negative detections \cite{free_lunch} and performing style transfer between the source and target domains \cite{umt_uda}. Recognising that pseudo-labels are inevitably unreliable, Chen \textit{et al.}\cite{prob_teacher} propose a probabilistic self-training framework that does not require confidence thresholds \cite{prob_ssod}. Instead, they implement an entropy focal loss that encourages the student to pay more attention to high certainty detections. While promising, these methods fail to leverage the rich contextual information available during robotic deployment and focus on shifts in the general appearance of images. Consequently, the potential for contextual and class distribution shift is largely ignored. Motivated by this, Xu \textit{et al.}\cite{categorical_regularisation} enforce consistency between the object detector and an image-level multilabel classifier, which is more robust to arbitrary changes in image background. However, this method was proposed for use in domain alignment instead of self-training. Cai \textit{et al.}\cite{object_relations} model the relationships between objects in a scene using a graph structure, and enforces consistency between student and teacher graphs. While incorporating contextual information improves robustness, the pseudo-labels and relational graphs generated using arbitrary thresholds remain unreliable. Furthermore, these methods do not consider the impact of class distribution shift on self-training. In response, we model image context to predict the class distribution of unlabelled data in novel deployment environments. This prior can be incorporated into self-training via ACT, which generates pseudo-labels to match the predicted class distribution \cite{labelmatch}. This method has shown promising results in SSOD, and we propose a series of changes to optimise it for domain adaptation. 

\vspace{-1em}
\subsection{Class Imbalance in Self-training}
It is well established that class imbalance leads to severe confirmation bias during self-training, as dominant classes are predicted with high confidence and subsequently reinforced \cite{unbiased1}. Solutions to this problem have been proposed for SSOD, such as the weighting of low probability classes \cite{unbiased1, prob_ssod} or ACT to align the distribution of pseudo-labels with a known class distribution \cite{labelmatch}. However, existing methods fail to address the class imbalance problem in the presence of domain shift.

\vspace{-1em}
\subsection{Class Distribution Shift in UDA-OD}
Existing UDA-OD benchmarks often focus on appearance changes due to factors such as weather, lighting conditions, or image corruptions \cite{dafrcnn}, but do not adequately assess realistic class distribution shifts. Ignoring class distribution shift is a critical limitation, as they are standard in open-world robotic deployment and can significantly impact the performance of object detection algorithms \cite{annotate_or_not, real_life_cl, wanderlust}. In particular, the Cityscapes \cite{cityscapes} to Foggy Cityscapes \cite{foggy_cityscapes} adaptation scenario contains no class distribution shift as the target domain is an augmented version of the source images \cite{dafrcnn}. Similarly, the cross camera adaptation and simulation to real scenarios only contain a single class \cite{dafrcnn}. The final benchmark studies adaptation from a small to large-scale dataset \cite{bdd}, reflecting the challenge of deploying a robot in a new environment with a broad array of conditions. We find that the shift in location produces significant class distribution shift, making it useful for this work. However, the impact of class distribution shift on UDA-OD is poorly evaluated using existing benchmarks. In turn, the benefits of explicitly addressing this problem during self-training are largely unexplored.

\vspace{-1em}
\subsection{Pre-trained Joint Vision and Language Models}
The abundance of captioned images has allowed the learning of transferable image representations using natural language supervision. One notable example is Contrastive Language-Image Pretraining (CLIP), which trains a model to predict the images and text that occur together using 400 million image-text pairs \cite{clip}. Because natural language descriptions are highly generalisable, CLIP can be easily transferred to new tasks and exhibits superior resistance to domain shift compared to supervised pre-training methods \cite{12-in-1, grounding_with_text, clip, ZegCLIP}. Language descriptions also contain a wealth of contextual information \cite{image_retrieval, learning_to_prompt, clip_optimal_representations}, making them useful for modelling the deployment environment of a robot. Recent work proposes using the similarity between an image and a series of text prompts produced by CLIP as a representation of image context \cite{ZegCLIP}. This representation is a strong predictor of object occurrence in an image, helping to improve semantic segmentation performance in open-world settings. In this work, we leverage the domain invariance and contextual detail of these image-text similarity scores to predict the class distribution of unlabelled data.

\section{Preliminaries}
\subsection{Problem Formulation}
In UDA-OD, a set of \(n_{l}\) labelled images \(D_{l}=\left\{X_{l},Y_{l} \right\}\) containing $n_{c}$ classes from the the source domain and \(n_{u}\) unlabelled images \(D_{u}=\left\{X_{u} \right\}\) from the target domain are provided. The goal of UDA-OD is to use both the labelled source and unlabelled target data to optimise performance of the object detector in the target domain. 

\vspace{-1em}
\subsection{Mean Teacher}
The Mean Teacher framework \cite{mean_teacher} encourages consistent predictions by a student and teacher model on augmented versions of the unlabelled data. This process leads the student and teacher to learn a solution from the source domain that is transferable to the target domain. An overview of this method is provided in Figure \ref{fig:MethodOverView}. The student and teacher models share the same network architecture, but different weights \(\theta^{s}\) and \(\theta^{t}\) respectively. For a given unlabelled sample \(x_{u}\), a weakly augmented version \(x_{u}^{t}\) is created for input into the teacher model\added{ by applying horizontal flipping and multi-scaling \cite{labelmatch}}. After non-maximum suppression (NMS), a confidence threshold is applied to the teacher's predictions \(f(x_{u}^{t},\theta^{t})\) to generate pseudo-labels \(y_{u}\). \replaced{Strong augmentations including color jittering, grayscale, gaussian blurring and cutout patches \cite{labelmatch} are}{A strong augmentation is} applied to create \(x_{u}^{s}\), which is used to generate the student's predictions \(f(x_{u}^{s},\theta^{s})\). In turn, an unsupervised loss can be calculated as:
\vspace{-0.25em}
\begin{equation} \label{eq:1}
\vspace{-0.25em}
\mathcal{L}_{u} = \mathcal{L}_{cls}(f(x_{u}^{s},\theta^{s}), y_{u}) + \mathcal{L}_{reg}(f(x_{u}^{s},\theta^{s}), y_{u})
\end{equation}
where \(\mathcal{L}_{cls}\) is the classification loss of the object detector, and \(\mathcal{L}_{reg}\) is the box regression loss. A standard supervised loss of the same form is calculated using a labelled sample \((x_{l},y_{l})\) from the source domain:
\vspace{-0.25em}
\begin{equation}
\vspace{-0.25em}
\mathcal{L}_{l} = \mathcal{L}_{cls}(\replaced{f(x_{l}^{s},\theta^{s})}{x_{l}},y_{l}) + \mathcal{L}_{reg}(\replaced{f(x_{l}^{s},\theta^{s})}{x_{l}},y_{l})
\end{equation}
The overall loss for a batch containing both labelled and unlabelled samples is then calculated as:
\vspace{-0.25em}
\begin{equation}
\vspace{-0.25em}
\mathcal{L} = \mathcal{L}_l + \lambda\mathcal{L}_{u}
\end{equation}
where \(\lambda\) is a hyperparameter to weight the unsupervised loss. 
This loss is used to update the student via gradient descent, and the exponential moving average (EMA) of the student's weights are used to update the teacher model. That is, after each training iteration \textit{t} the weights of the teacher model are calculated as:
\vspace{-0.25em}
\begin{equation}
\vspace{-0.25em}
\theta_{t}^{t} = \alpha\theta_{t-1}^t + (1-\alpha)\theta_{s}
\end{equation}
where \(\alpha\) controls the stability of the teacher model. 
\vspace{-1em}
\subsection{Adaptive Label Distribution-aware Confidence Thresholding (ACT)}
ACT aims to select confidence thresholds for Mean Teacher to overcome the class imbalance problem. It does so by aligning the class distribution of the pseudo-labels with that of the labelled data \cite{labelmatch}. The method firstly calculates the class ratio of the labelled data \(\bm{r^{l}} = [r^{l}_{1},...,r^{l}_{n_{c}}]\) and the number of objects per image \(n^{l}_{o}\). Using this class distribution prior, the expected number of objects in the unlabelled data for each class $c$ is calculated as:
\vspace{-0.25em}
\begin{equation}
\vspace{-0.25em}
n_{c}^{u} = r^{l}_{c} \cdot n^{l}_{o} \cdot n_{u}
\end{equation}
The confidence threshold for class $c$ is then selected such that \(n_{c}^{u}\) pseudo-labels are generated. That is:
\vspace{-0.25em}
\begin{equation}
\vspace{-0.25em}
t_{c} = \bm{p_{c}}^{sort}[n_{c}^{u}]
\end{equation}
where $\bm{p_{c}^{sort}}$ is a list of the confidence scores predicted for class $c$ by the teacher, sorted in descending order. ACT can be used with any self-training approach based on Mean Teacher, but specific changes to the standard implementation are recommended by the authors \cite{labelmatch}. A proportion of the pseudo-labels are defined as reliable and used to calculate the hard classification and box regression losses defined in (\ref{eq:1}). To deal with the noise present in the unreliable pseudo-labels, a soft classification loss is calculated using the cross-entropy between student and teacher classification predictions \cite{labelmatch}. This results in a reformulation of (\ref{eq:1}) as:
\vspace{-0.25em}
\begin{equation}
\vspace{-0.25em}
\begin{split}
\mathcal{L}_{u} = \mathcal{L}_{cls}(f(x_{u}^{s},\theta^{s}), y_{ur}) + \mathcal{L}_{reg}(f(x_{u}^{s},\theta^{s}), y_{ur}) \\ + \widehat{\mathcal{L}}_{cls}(f(x_{u}^{s},\theta^{s}), y_{uu})
\end{split}
\end{equation}
where $y_{ur}$ and $y_{uu}$ refer to the reliable and unreliable pseudo-labels, and $\widehat{\mathcal{L}}_{cls}$ denotes the soft classification loss.

\vspace{-1em}
\subsection{Contrasting Language-Image Pretraining (CLIP)}
Given a batch of $n$ images $X_{i}$ and associated texts $X_{t}$, CLIP is trained to predict which possible image-text pairs $X_{i} \times X_{t}$ actually occurred. Images and texts are input to an image encoder $I$ and text encoder $T$ to extract their respective embeddings. These embeddings are then projected into a shared, multi-modal representation space with learnt projection matrices $W_{i}$ and $W_{t}$. L2-normalisation is then applied to extract the final multi-modal embeddings $Z_{i} \in \mathbb{R}^{n \times d}$ and $Z_{t}  \in \mathbb{R}^{n \times d}$:
\vspace{-0.25em}
\begin{equation} \label{eq8}
\vspace{-0.25em}
Z_{i} = \left\| I(X_{i}) \cdot W_{i} \right\|_{2}
\end{equation}
\begin{equation} \label{eq9}
Z_{t} = \left\| T(X_{t}) \cdot W_{t} \right\|_{2}
\vspace{-0.25em}
\end{equation}
The cosine similarity of the shared embeddings is then calculated, and multiplied by a learnt temperature parameter $t$ to calculate similarity scores $S \in \mathbb{R}^{n \times n}$ for each text and image pair in the batch:
\begin{equation} \label{eq10}
S = (Z_{i} \cdot Z_{t}^T)*exp(t)
\end{equation}
During training, a cross-entropy loss is applied to the similarity scores to encourage similarity between true image-text pairs and discourage similarity between false pairs. Our work uses the pretrained model to generate similarity scores between images and image classification labels.
\vspace{-1em}
\section{Method}
\begin{figure*}[!t]
    \centering
    \vspace{-0em}
    \includegraphics[width=1.99\columnwidth, height=0.16\textheight]{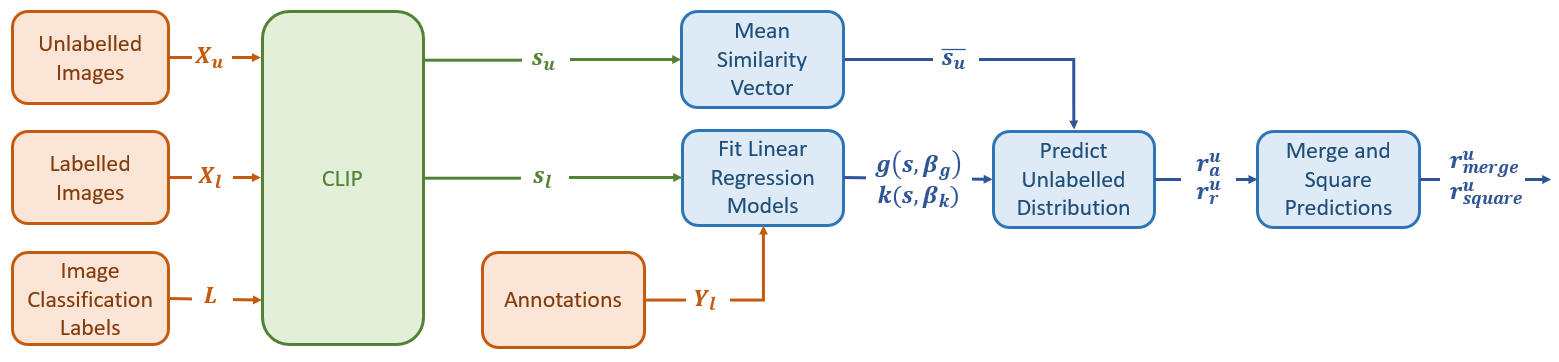}
    \vspace{-0.5em}
    \caption{Our proposed method for predicting the class ratio of the unlabelled data. CLIP is used to calculate the similarity between the labelled images $X_{l}$ and a series of \replaced{natural language}{image classification} labels $L$ of the form ``a photo of class $c$". Using the labelled similarity vector $\bm{s_{l}}$ as a domain invariant representation of semantic context, two linear regression models are fit to predict the number of instances $g(\bm{s}, \beta_{g})$ and the class ratio $k(\bm{s}, \beta_{k})$ in each labelled image. To make a prediction for the class ratio of the unlabelled images $X_{u}$, CLIP is used to extract the similarity vectors $\bm{s_{u}}$. The mean similarity vector $\bm{\overline{s_{u}}}$ is then calculated and input to the linear regression models to generate two distinct predictions for the class ratio of the entire unlabelled dataset. These predictions are merged by calculating the geometric mean, and the relative change in class ratio squared to account for persistent underestimation.}
    \label{fig:class_ratio_prediction}
    \vspace{-1em}
\end{figure*}

\subsection{Class Distribution Prediction}
We aim to predict the class ratio of the unlabelled data $\bm{r^{u}}\in \mathbb{R}^{n_{c}}$, such that it can be used as prior for performing ACT. An overview of this approach can be found in Figure \ref{fig:class_ratio_prediction}. We start by using CLIP to generate a representation of the semantic context of each image. To do so, we calculate the similarity between the labelled images $X_{l}$ and a series of  \added{natural language prompts $L$. The performance of vision-language models on downstream tasks is sensitive to the format of natural language labels, leading to the development of prompt optimisation techniques \cite{learning_to_prompt, unsupervised_prompt_learning}. We implement a simple approach to generate $L$ where each class label $c$ is converted to a prompt of the form ``a photo of $c$" \cite{clip}.}\deleted{image recognition labels $L$. Following the use of CLIP for zero-shot image recognition \cite{clip}, the set of labels is of the form ``a photo of $c$" for each class $c$ in the labelled dataset.} For each labelled image, this results in a vector of similarity scores $\bm{s_{l}}\in \mathbb{R}^{n_{c}}$ that characterise how the image relates to the text prompts \cite{ZegCLIP}. As the natural language descriptions used to generate the similarity vectors are resistant to domain shift \cite{12-in-1, grounding_with_text}, this results in a contextual representation that is consistent across the source and target data.

We investigate two linear regression models, g and k, for predicting the class ratio given the similarity vector $\bm{s}$ as a domain invariant input. The absolute model $g(\bm{s}, \beta_{g})$ is optimised using the labelled data to predict the absolute number of instances of each class in an image. By normalising the output of $g(\bm{s}, \beta_{g})$ to sum to 1, this model can be used to predict the class ratio. The relative model $k(\bm{s}, \beta_{k})$ is optimised using the labelled data to directly predict the ratio of each class in an image. The relative model does not consider how many objects occur in an image, instead learning how likely a class is to occur relative to other classes. 

To make a prediction for the class distribution of the unlabelled dataset, we use CLIP to extract the similarity vector $s_{u}$ for each unlabelled image. We then calculate the mean similarity vector across the unlabelled dataset $\bm{\overline{s_{u}}}\in\mathbb{R}^{n_{c}}$ and input it to the linear regression models:
\vspace{-0.25em}
\begin{equation} 
\vspace{-0.25em}
\bm{r^{u}_{a}}  = \frac{g(\bm{\overline{s_{u}}}, \beta_{g})}{\left|g(\bm{\overline{s_{u}}}, \beta_{g})\right|}
\end{equation}
\begin{equation} 
\bm{r^{u}_{r}}  = k(\bm{\overline{s_{u}}}, \beta_{k})
\end{equation}

\replaced{The relative and absolute models produce distinct predictions $\bm{r^{u}_{a}}\in\mathbb{R}^{n_{c}}$ and $\bm{r^{u}_{r}}\in\mathbb{R}^{n_{c}}$ for the class ratio, either of which can be used in our framework. However, neither prediction is consistently more accurate than the other, and the optimal prediction to use for a specific scenario cannot be determined without labelled data. To ensure consistent performance across all scenarios, we therefore average the output of the two models.}{We found that combining the prediction from the absolute and relative models improved the accuracy.} To do so, we propose using the geometric mean, as it is more appropriate than the arithmetic mean for summarising the central tendency of ratios. We term the element-wise geometric mean of the class ratio predictions $\bm{r^{u}_{merge}}\in\mathbb{R}^{n_{c}}$ as the merged prediction:
\begin{equation} 
\bm{r^{u}_{merge}}  = \sqrt{\bm{r^{u}_{a}}\cdot \bm{r^{u}_{r}}}
\end{equation}

Secondly, we found that the merged prediction persistently underestimates the shift from the labelled class ratio $\bm{r^{l}}$ to the unlabelled class ratio $\bm{r^{u}}$. \deleted{That is, if the merged prediction states the class ratio for cars should increase from 0.5 in the labelled data to 0.6 in the unlabelled data, the true value is generally greater than 0.6.} A heuristic solution to this underestimation problem is to square the relative change in the class ratio predicted by $\bm{r^{u}_{merge}}$. Formally, we express $\bm{r^{u}_{merge}}$ relative to $\bm{r^{l}}$ and calculate the squared prediction $\bm{r^{u}_{square}}$ for the class ratio as: 
\vspace{-0.25em}
\begin{equation} 
\vspace{-0.25em}
\bm{r^{u}_{square}}  = \bm{r^{l}}\left(\frac{\bm{r^{u}_{merge}}}{\bm{r^{l}}}\right)^{2} = \frac{\bm{r^{u}_{a}}\cdot\bm{ r^{u}_{r}}}{\bm{r^{l}}}
\end{equation}
Finally, we apply L1 normalisation to $\bm{r^{u}_{square}}$ so the distribution sums to 1.

\vspace{-1em}
\subsection{Dynamic Selection of Pseudo-labels per Image}
Even with a perfectly accurate class distribution, we find that under significant domain shift the pseudo-labels generated by the teacher remain unreliable early in self-training. Thus, instead of estimating the average number of objects per image in the unlabelled data, we set this value to target a mean confidence score $\tau$ across the generated pseudo-labels. This accounts for the initial poor performance of the teacher in the target domain, and allows the number of pseudo-labels per image to increase as the model becomes more confident. We use Algorithm \ref{alg:alg1} to calculate the number of objects per image required to deliver a mean pseudo-label confidence score of $\tau$. In this algorithm, the average number of objects per image $n_{o}\in\mathbb{R}$ is iteratively increased by $\Delta n_{o}$. \added{At each step, pseudo-labels $y_{u}$ are generated for the entire unlabelled dataset by aligning them with the predicted class ratio $\bm{r^{u}}$ and the current number of objects per image $n_{o}$. The mean confidence score $p_{mean}$ is then calculated across the pseudo-labels, and the loop terminates when this value falls below the target value $\tau$.}\deleted{At each step, ACT is applied using the predicted class ratio $\bm{r^{u}}$ to generate pseudo-labels $y_{u}$, and the mean confidence score $p_{mean}$ is calculated. The loop terminates and the required number of objects per image is returned when $p_{mean}$ falls below the target value $\tau$.}

\begin{algorithm}[H]
\caption{Dynamic Selection of Objects per Image}\label{alg:alg1}
\begin{algorithmic}
\STATE
\STATE $\bm{r^{u}}$ = \textsc{PredictClassRatio}($D_{l}$, $D_{u}$)
\STATE $n_{o} = 0$
\STATE \textbf{while} True \textbf{do}
\STATE \hspace{0.5cm} $n_{o} = n_{o} + \Delta n_{o}$
\STATE \hspace{0.5cm} $y_{u}$ = \textsc{ACT}($\bm{r^{u}}$, $n_{o}$)
\STATE \hspace{0.5cm} $p_{mean}$ = \textsc{MeanConfidenceScore}($y_{u}$)
\STATE \hspace{0.5cm} \textbf{if} $p_{mean}<\tau$ \textbf{then}
\STATE \hspace{1.0cm} \textbf{return} $n_{o} - \Delta n_{o}$
\end{algorithmic}
\label{alg1}
\end{algorithm}

\vspace{-1.5em}
\section{Experiments}
\subsection{Experimental Settings}
\textbf{Scenarios}. We validate our proposed method using two scenarios previously used for UDA-OD, and one novel scenario that captures a more extreme class distribution shift. 
\begin{itemize}
\item{Adaptation from small to large dataset (C2B): as in previous work \cite{prob_teacher, free_lunch, categorical_regularisation}, we utilise Cityscapes \cite{cityscapes} as a small source dataset and BDD100k daytime \cite{bdd} images as the large and diverse target domain. \added{The shift in location in this scenario produces significant class distribution shift.}} 
\item{Adaptation from normal to foggy weather (C2F): this commonly studied scenario uses Cityscapes \cite{cityscapes} as the source domain and Foggy Cityscapes \cite{foggy_cityscapes} as the target domain. To compare with the optimal performance of the current state of the art \cite{prob_teacher}, we use all three levels of fog.}
\item{Adaptation from daytime to night (C2N): we propose a novel benchmark to \deleted{further} explore the impact of \added{more extreme} class distribution shift on UDA-OD. We utilise Cityscapes \cite{cityscapes} as the source domain and BDD100k night \cite{bdd} as the target domain.}
\end{itemize}

\textbf{Network Architecture.} \replaced{A standard object detection architecture is used across all UDA-OD works to enable fair comparison between methods \cite{dafrcnn}. Following this trend,}{Following previous works on domain adaptation \cite{dafrcnn},} we use Faster-RCNN \cite{faster_rcnn} as the detector architecture with VGG-16 \cite{vgg} pre-trained on ImageNet \cite{imagenet} as the backbone. \added{The inference rate of this network is quoted at 5Hz on a Nvidia K40 GPU \cite{faster_rcnn}, while our experiments return 20Hz on a Nvidia GeForce RTX 3090. However, our method could be easily extended to novel architectures, making it useful for diverse robotic applications \cite{labelmatch}.}

\textbf{Implementation Details.} We follow the default implementation provided in \cite{labelmatch} to test our method, which leverages MMDetection \cite{mmdet}. Thus, we use a batch size of 16 for the labelled and unlabelled data and a learning rate of 0.016. The optimiser used is SGD, with a momentum rate of 0.9 and a weight decay of 0.0001. For all experiments, the model is pre-trained using source data for 4,000 iterations before beginning self-training. For the C2B and C2N scenarios, we implement 42,000 iterations of self-training. Due to slower convergence on C2F, we self-train for 72,000 iterations. Lastly, we use a target mean confidence score $\tau$ of 0.6 and a step size $\Delta n_{o}$ of 0.1 for selecting the number of objects per image. 

\textbf{Evaluation.} Standard mean average precision (mAP) at an IOU threshold of 0.5 is used to compare performance with that reported by existing works. \added{In additional to implementing our approach, we conduct the following evaluations:}
\begin{itemize}
    \item \added{For comparison on the novel C2N scenario, we implement Probabilistic Teacher \cite{prob_teacher} as per the released code and report the peak performance recorded across 24,000 iterations. This method is the current state of the art on the C2B and C2F scenarios, and thus provides a strong baseline for evaluating our method under more extreme class distribution shift.}
    \item \added{We evaluate the original ACT \cite{labelmatch} on all scenarios using the same implementation as our method.}
    \item \added{Lastly, for the C2N scenario, we evaluate the model trained for 48,000 iterations on the source data, and target data only.}
\end{itemize}
\deleted{Standard mean average precision (mAP) at an IOU threshold of 0.5 is used to compare to existing methods. For comparison on the novel C2N scenario, we implement Probabilistic Teacher \cite{prob_teacher} as per the released code and report the peak performance recorded across 24,000 iterations. This method is the current state of the art on the C2B and C2F scenarios, and thus provides a strong baseline for evaluating our method under more extreme class distribution shift. We also implement the original ACT \cite{labelmatch} on all scenarios using the same implementation as our method. Lastly, for the C2N scenario, we evaluate the model trained for 48,000 iterations on the source data, and target data only.}

\renewcommand\arraystretch{1}
\begin{table}[t]
	\centering
	\caption{Results of adaptation from small to large-scale dataset (C2B) and Cityscapes to BDD100k night (C2N) scenario. \added{ ``Source only'' and  ``Target only'' refer to the models trained by only using labelled source data and labelled target data.}\deleted{See Table~\ref{tab:c2f_results} for an explanation of naming conventions.}}
        \vspace{-0.5em}
        \begin{threeparttable}
    \resizebox{.99\columnwidth}{!}{
	\begin{tabular}{l | c | c | c}
		\toprule
		Methods & Reference & C2B (mAP) & C2N (mAP) \\
		\hline
		Source only \cite{prob_teacher} & - & 20.6 & 3.8 \\
		\cline{1-4}
		ICR-CCR \cite{categorical_regularisation} & CVPR'20 & 26.9 & -  \\
		SFOD \cite{free_lunch} & AAAI'21  & 29.0 & - \\
		PT \cite{prob_teacher} & ICML'22  &  34.9 & 19.5\\	
		\cline{1-4}
		LabelMatch \cite{labelmatch} & - & 30.3 & 11.1\\
		Ours  & - &  \textbf{39.2} & \textbf{24.2}\\
		\cline{1-4}
		Target only \cite{prob_teacher} & -  & 51.7 & 32.7 \\
	 \hline
	\end{tabular}
    }
	\end{threeparttable}
	\label{tab:c2b_results}
	\vspace{-1em}
\end{table}

\vspace{-1em}
\subsection{Performance Comparison}
Our method matches or outperforms existing approaches across all three scenarios (Table~\ref{tab:c2b_results}, \ref{tab:c2f_results}). Strong performance occurs on scenarios with more extreme class distribution shift, a case that has been largely overlooked in existing benchmarks. We outperform the strongest baseline Probabilistic Teacher \cite{prob_teacher} by 4.3 mAP on the previously studied C2B scenario (Table~\ref{tab:c2b_results}). We further return a 4.7 mAP improvement on the novel C2N scenario (Table~\ref{tab:c2b_results}), which is more challenging due to extreme class distribution shift (Figure~\ref{fig:driving_class_pred}) and the discrete appearance change from daytime to night. A clear improvement is also noted on both scenarios relative to using the original ACT \cite{labelmatch}. These results highlight that by modelling shifts in a robot's deployment context, pseudo-label reliability can be improved in Mean Teacher. Furthermore, it emphasises that class distribution shift must be considered when designing UDA-OD methods for robotics applications.

Interestingly, our approach remains competitive with existing methods on the normal to foggy scenario (C2F) (Table~\ref{tab:c2f_results}), where there is a large change in image appearance but no class distribution shift. The average performance across five trials of our method is identical to the optimal performance returned by Probabilistic Teacher. The benefit of our approach is \deleted{most} prominent in low probability classes such as train, truck and bus, indicating that we are effectively mitigating the class imbalance problem. Furthermore, we return a 4.1 mAP improvement on this scenario relative to the original ACT, emphasising the benefit of accounting for pseudo-label reliability by dynamically updating the number of objects per image.

\renewcommand\arraystretch{1.00}
\begin{table*}[th]
\vspace{1em}
	\centering
	\vskip -0.1in
	\caption{Results of adaptation from normal to foggy weather (C2F) \cite{prob_teacher}. Due to variability in training and similar performance to Probabilistic Teacher, we run our method five times and report the mean mAP and standard deviation.\deleted{ ``Source only'' and  ``Target only'' refer to the models trained by only using labelled source data and labelled target data respectively, as reported in \cite{prob_teacher}.} ``0.01'', ``0.02'' and ``ALL'' in the column of ``split'' represent the level of fog included in the target domain. \added{See Table~\ref{tab:c2b_results} for details of naming conventions.}}
        \vspace{-0.5em}
	\begin{threeparttable}
	\resizebox{.99\textwidth}{!}{
	\begin{tabular}{l | c | c | cccccccc | c}
		\toprule
		Methods & Split & Reference & truck  & car & rider & person & train & motor & bicycle & bus  & mAP\\
		\hline
		Source only \cite{prob_teacher} & ALL  & - & 12.1 & 40.4 & 33.4  & 27.9 & 10.1 & 20.7 & 30.9 & 23.2 & 24.8 \\
		\hline
		MTOR \cite{object_relations} & 0.02  & CVPR'19 & 21.9 & 44.0 & 41.4 & 30.6 & 40.2 & 31.7 & 33.2 & 43.4 & 35.1 \\
		\hline
		SW \cite{strong_weak_alignment} & 0.02  & CVPR'19  & 24.5 & 43.5 & 42.3 & 29.9 & 32.6 & 30.0 & 35.3 & 32.6 & 34.3 \\
		\hline
		DM \cite{diversify_and_match} & UN & CVPR'19 & 27.2 & 40.5 & 40.5 & 30.8 & 34.5 & 28.4 & 32.3 & 38.4 & 34.6 \\
		\hline
		PDA \cite{progressive_adaptation} & ALL & WACV'20 & 24.3 & 54.4 & 45.5 & 36.0 & 25.8 & 29.1 & 35.9 & 44.1  & 36.9\\
		\hline
		GPA \cite{gpa} & 0.01 & CVPR'20 & 24.7 & 54.1 & 46.7 & 32.9 & 41.1 & 32.4 & 38.7 & 45.7  & 39.5 \\
		\hline
		ATF \cite{ATF} & UN & ECCV'20 & 23.7 & 50.0 & 47.0 & 34.6 & 38.7 & 33.4 & 38.8 & 43.3 & 38.7  \\
		\hline
		HTCN \cite{HTCN} & 0.02  & CVPR'20  & 31.6 & 47.9 & 47.5 & 33.2 & 40.9 & 32.3 & 37.1 & 47.4 & 39.8 \\
		\hline
		ICR-CCR \cite{categorical_regularisation} & ALL  & CVPR'20 & 27.2 & 49.2 & 43.8 & 32.9 & 36.4 & 30.3 & 34.6 & 36.4 & 37.4 \\
		\hline
		CF \cite{coarse} & UN & CVPR'20 & 30.8 & 52.1 & 46.9 & 34.0 & 29.9 & 34.7 & 37.4 & 43.2  & 38.6 \\
		\hline
		iFAN \cite{ifan} & UN & AAAI'20 & 27.9 & 48.5 & 40.0 & 32.6 & 31.7 & 22.8 & 33.0 & 45.5  & 35.3 \\
		\hline
		SFOD \cite{free_lunch} & ALL & AAAI'21 & 25.5 & 44.5 & 40.7 & 33.2 & 22.2 & 28.4 & 34.1 & 39.0  & 33.5 \\
		\hline
		MeGA \cite{mega} & UN  & CVPR'21 & 25.4 & 52.4 & 49.0 & 37.7 & \textbf{46.9} & 34.5 & 39.0 & 49.2  & 41.8\\
		\hline
		UMT  \cite{umt_uda} & 0.02 & CVPR'21 & 34.1 & 48.6 & 46.7 & 33.0 & 46.8 & 30.4 & 37.3 & 56.5 & 41.7  \\
		\hline
		PT \cite{prob_teacher} & ALL & ICML'22 & 33.4 & \textbf{63.4} & \textbf{52.4} & \textbf{43.2} & 37.8 & \textbf{41.3} & \textbf{48.7} & 56.6  & \textbf{47.1} \\
		\hline
		LabelMatch \cite{labelmatch} & ALL & - & 33.4 & 61.3 & 49.8 & 38.7 & 27.4 & 32.7 & 44.5 & 55.8 & 43.0 \\
		\hline
		Ours & ALL & - & \textbf{36.3$\pm$3.8} & 61.2$\pm$0.7 & 49.2$\pm$2.3 & 41.9$\pm$0.9 & 44.2$\pm$5.9 & 37.3$\pm$2.1 & 47.7$\pm$0.5 & \textbf{59.2$\pm$0.2} & \textbf{47.1$\pm$1.3}  \\
		\hline
		Target only \cite{prob_teacher} & ALL & - & 32.6 & 61.6 & 49.1 & 41.2 & 49.0 & 37.9 & 42.4  &  56.6  & 46.3  \\
	 		\hline
	\end{tabular}
        }
	\end{threeparttable}
	\label{tab:c2f_results}
	\vspace{-1.5em}
\end{table*}

\renewcommand\arraystretch{1.0}
\begin{table}[t]
	\centering
        \vspace{-0.5em}
	\caption{Final validation performance on the C2B scenario when using alternative class ratios as the prior and different approaches to setting the number of pseudo-labels per image.}
        \vspace{-0.5em}
	\begin{threeparttable}
	\begin{tabular}{l | c | c }
		\toprule
		Obj. per Img. & Class Ratio & mAP \\
		\cline{1-3}
		Labelled Data & Labelled Data & 30.3\\
		\cline{1-3}
		Dynamic (Ours) & Labelled Data & 35.4\\
		\cline{1-3}
		Dynamic (Ours) & Predicted (Ours) &  39.2\\
		\cline{1-3}
		Dynamic (Ours) & Unlabelled Data &  40.2\\
	 		\hline
	\end{tabular}
	\end{threeparttable}
	\label{tab:prior_ablation}
\end{table}

\renewcommand\arraystretch{1.0}
\begin{table}[t]
	\centering
	\vskip -0.1in
	\caption{Sensitivity of our approach on the C2B scenario to different settings of the target mean confidence threshold $\tau$.}
        \vspace{-0.5em}
	\begin{threeparttable}
	\begin{tabular}{l | c | c | c | c | c  }
		\toprule
		$\tau$ & 0.4 & 0.5 & 0.6 & 0.7 & 0.8 \\
		\cline{1-6}
		mAP & 29.9 & 36.1 & 39.2 &  40.5 &  23.3\\
			\hline
	\end{tabular}
	\end{threeparttable}
	\label{tab:target_confidence_sensitivity}
        \vspace{-1em}
\end{table}

\vspace{-0.5em}
\subsection{Impact on Pseudo-label Reliability}
\begin{figure*}[!t]
    \centering
    \includegraphics[width=1.99\columnwidth, height=0.15\textheight]{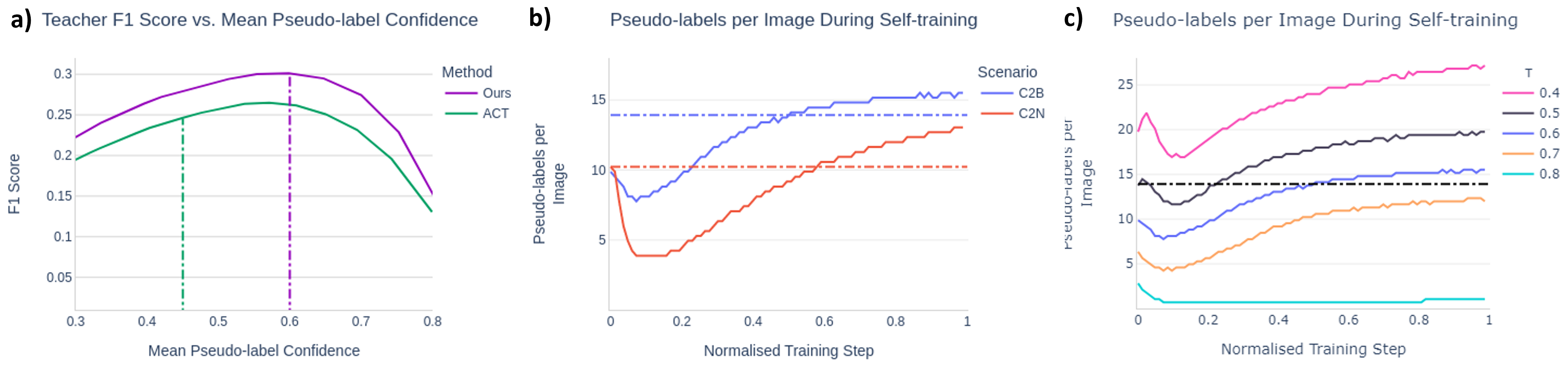}
    \vspace{-1em}
    \caption{a) The F1 score of pseudo-labels generated by the teacher model after 1000 iterations of self-training using different class distribution priors. The solid lines highlight how F1 score varies with respect to mean pseudo-label confidence when using different class ratios as prior. The dotted lines show the mean confidence score selected using each method. The pseudo-labels generated by our method (purple) have an F1 score of 0.3, while those generated using ACT (green) return 0.24. b) Evolution of the number of pseudo-labels per image while training the proposed method on the C2N and C2B scenarios. \added{The dotted lines show the true number of objects per image in the target dataset for each scenario. }c) \replaced{Evolution of the number of pseudo-labels per image while training on the C2B scenario with different settings of the mean pseudo-label confidence $\tau$. The black dotted line shows the true number of objects per image.}{Evolution of reliable pseudo-label accuracy while training the proposed method on the C2N and C2B scenarios.}}
    \vspace{-1.5em}
    \label{fig:pseudo-label}
\end{figure*}

To further explore the proposed approach, we assess the pseudo-label reliability generated using alternative threshold selection techniques on the C2B scenario. To do so, we take the teacher model after 1000 iterations of self-training and calculate the F1 score of the generated pseudo-labels when different class ratios and number of objects per image are used (Figure~\ref{fig:pseudo-label}a). This plot firstly highlights the benefit of using an accurate class distribution prediction to inform pseudo-label selection. When the predicted class ratio is used, there is a clear shift to higher F1 scores relative to using the naive prior provided by the labelled data. We further see that targeting a mean confidence score of 0.6 results in an F1 score that is close to the peak. Consequently, the proposed approach results in an improvement in the F1 score of 0.06 relative to the original ACT. We also plot how the selected number of objects per image evolves through training for the C2B and C2N scenarios, highlighting how the proposed method responds to the variable reliability of the teacher model (Figure~\ref{fig:pseudo-label}b-c). As the pseudo-label accuracy improves with training, the number of objects per image is dynamically increased. We also see that for the more challenging C2N scenario, the number of pseudo-labels per image is lower to account for the poorer performance of the teacher in the target domain.

\vspace{-1em}
\subsection{Class Distribution Prediction}
\begin{figure}[!t]
    \centering
    \includegraphics[width=0.99\columnwidth, height = 0.18\textheight]{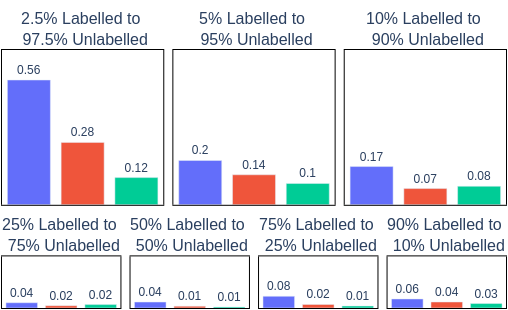}
    \vspace{-1.5em}
    \caption{KL divergence of alternative class ratio prediction methods for different temporal splits of the Wanderlust dataset. Each subplot shows an adaptation scenario, defined by the proportion of the data considered labelled. \replaced{See Figure 4 for a description of the methods assessed.}{The blue bar represents the KL divergence between the labelled and unlabelled class ratios. The red bar shows the KL-divergence between our merged prediction and the unlabelled class ratio. The green bar shows the KL-divergence of the squared prediction and the unlabelled class ratio.}}
    \vspace{-1.5em}
    \label{fig:oak_class_pred}
\end{figure}

\begin{figure}[!t]
    \centering
    \includegraphics[width=0.99\columnwidth, height = 0.25\textheight]{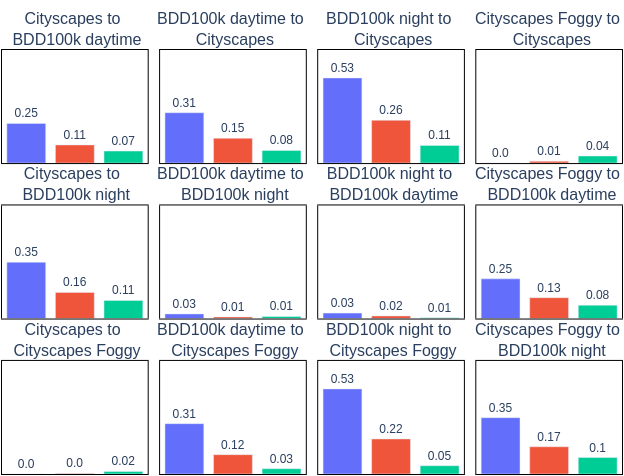}
    \vspace{-1.5em}
    \caption{KL divergence of alternative class ratio prediction methods from the true distribution on different combinations of the driving datasets. The title of each subplot shows the adaptation scenario. \replaced{The blue bar represents the KL divergence between the labelled and unlabelled class ratio, with a larger value corresponding to scenarios with greater class distribution shift. The red and green bars show the KL-divergence between the unlabelled class ratio and the proposed merged and squared predictions respectively. Smaller values correspond to a more accurate prediction of the class ratio.}{The blue bar represents the KL divergence between the labelled and unlabelled class ratio. The red bar shows the KL-divergence between our merged prediction and the unlabelled class ratio. The green bar shows the KL-divergence of the squared prediction and the unlabelled class ratio.}}
    \vspace{-1.5em}
    \label{fig:driving_class_pred}
\end{figure}

We further investigate the accuracy of the proposed class ratio prediction method to assess if it is robust to different forms of domain shift. To do so, we utilise the four driving datasets used to assess UDA-OD performance; Cityscapes \cite{cityscapes}, foggy Cityscapes \cite{foggy_cityscapes}, BDD100k daytime and BDD100k night \cite{bdd}. The permutation of these datasets into labelled and unlabelled pairs leads to 12 scenarios for predicting class distribution shift (Figure~\ref{fig:driving_class_pred}). Additionally, we use the Wanderlust dataset \cite{wanderlust} to assess realistic class distribution shift encountered across time by an egocentric agent. This dataset contains object annotations of an egocentric videostream collected by a graduate student over nine months of their life. The seven most common classes of bicycle, bus, car, chair, dining table, person and potted plant are considered. Temporally coherent splits of the training data are created by splitting it at the 2.5\%, 5\%, 10\%, 25\%. 50\%, 75\% and 90\% mark of the datastream. We use the labels for all data before the split, resulting in 7 scenarios for assessing class distribution prediction (Figure~\ref{fig:oak_class_pred}).

For each scenario, we report the Kullback-Leibler (KL) divergence between the labelled class ratio, our merged and squared predictions, and the true class ratio. Generally, the squared prediction is much more accurate than using the labelled data alone, reducing KL divergence by 3-4 fold. On the driving scenarios, our method is robust to large and small class distribution shifts and significant appearance changes due to foggy weather and night driving (Figure~\ref{fig:driving_class_pred}). This emphasises that the representation of semantic context used to predict the class ratio is sufficiently resistant to domain shift. On the Wanderlust scenarios, our prediction remains accurate across a variable labelling budget (Figure~\ref{fig:oak_class_pred}). 

 \replaced{We also investigate the regression models to ascertain why the squared prediction is consistently more accurate than the merged prediction (Figure~\ref{fig:driving_class_pred}, \ref{fig:oak_class_pred}). We find that the intercept coefficients of these models vary significantly across datasets. This implies that the class ratio is influenced by exogenous factors not captured by the CLIP similarity scores. As a result, a model fit to the labelled data will consistently be in error when making predictions on the shifted, unlabelled data. However, this intercept error is strongly correlated with the class distribution shift between the datasets ($r = 0.98$ for the driving scenarios in Figure~\ref{fig:driving_class_pred}). Thus, by predicting the relative change in class ratio between datasets, we are indirectly estimating the intercept error. Scaling up the predicted change in class ratio therefore improves accuracy by incorporating this prediction of the intercept error into our method. In future, prompt learning techniques \cite{learning_to_prompt, unsupervised_prompt_learning} could be investigated to make the CLIP similarity scores more expressive of the class ratio, avoiding the need to heuristically account for the intercept error.}{These results also show that the squared prediction is consistently more accurate than the merged prediction. Evidently, the proposed linear regression models are persistently underestimating the shift in class distribution from the labelled to unlabelled data. That is, the \textit{direction} of the shift predicted by these models is correct, but the \textit{magnitude} is too small. In future work, a more robust solution to this problem than our simple heuristic of squaring the predicted shift could be investigated. In particular, non-linear models may better capture the relationship between the CLIP similarity scores and the class ratio. To use such techniques, discretisation of the class ratio and the subsequent use of classification techniques may be required to overcome noise in the per-image class ratio. Image augmentation could also be applied before extracting CLIP embeddings to increase robustness to domain shift.}

\vspace{-1em}
\subsection{Ablations}
Using the C2B scenario, we study the independent effect of using a more accurate class ratio and the dynamic adjustment of the number of objects per image. We first introduce the dynamic adjustment of the number of pseudo-labels per image, but continue using the labelled class distribution prior. The resulting 5.1 mAP improvement relative to the original ACT \cite{labelmatch} can therefore be accredited to using the mean confidence of the teacher to select the number of pseudo-labels per image (Table~\ref{tab:prior_ablation}). When the predicted prior is introduced instead of the labelled class ratio, a further improvement of 3.8 mAP results. Lastly, we test the performance of our method if the true class ratio of the unlabelled data was used as the prior, finding that it leads to only a minor increase in performance, from 39.2 to 40.2 mAP (Table~\ref{tab:prior_ablation}).

We also study the sensitivity of our approach under different settings of the target mean pseudo-label confidence $\tau$. For values between 0.5 and 0.7, our approach maintains performance superior to the strongest baseline (Table~\ref{tab:target_confidence_sensitivity}). \replaced{These values perform optimally as they lead the number of pseudo-labels per image to converge to the true number of objects per image in the target dataset (Figure~\ref{fig:pseudo-label}b-c). We therefore recommend setting $\tau=0.6$ as a starting point on novel scenarios, before confirming that the number of pseudo-labels per image is converging to an appropriate value.}{However, a  drop in performance is noted when $\tau$ is increased to 0.8, at which point the number of pseudo-labels per image fails to increase during training. To account for this, we use the more conservative value of 0.6 for our experiments.} In future work, merging class distribution information with probabilistic self-training approaches \cite{prob_ssod, prob_teacher} may reduce sensitivity to key parameters.

\vspace{-0.5em}
\section{Conclusion}
In this work, we propose a framework for explicitly considering class distribution shift to improve the reliability of pseudo-labels in self-training. The resulting method improves upon existing baselines on a range of UDA-OD scenarios, showing strong performance when facing realistic contextual and class distribution shifts. The results motivate further research into how a robot's deployment environment can be modelled to improve pseudo-label reliability in Mean Teacher. Furthermore, we show that class distribution shift must be considered when implementing UDA-OD methods for robotics applications. Lastly, the weak supervision provided by the class distribution prior may benefit other tasks crucial to reliable robotic deployment, such as run-time monitoring or out-of-distribution detection.

\bibliographystyle{IEEEtran} 
\bibliography{IEEEabrv,main} 
\end{document}